# Capability-Guided Compression: Toward Interpretability-Aware Budget Allocation for Large Language Models

**Rishaank Gupta**

*Independent Researcher*

March 2026

## Abstract

Large language model compression has made substantial progress through pruning, quantization, and low-rank decomposition, yet a fundamental limitation persists across all existing methods: compression budgets are allocated without any representation of what individual model components functionally encode. We term this the capability-blind compression problem and argue it is a root cause of two well-documented failures—the insensitivity of perplexity-based evaluation to reasoning capability loss, and the abrupt phase transitions in model performance recently characterized by Ma et al. (2026). We propose Capability-Guided Compression (CGC), a framework that addresses this by using Sparse Autoencoder (SAE)-derived capability density maps to allocate differential compression budgets across transformer components. Capability density is a formally defined scalar measure combining the feature breadth, activation entropy, and cross-input consistency of a component's SAE feature activation distribution. We prove theoretically that components with higher capability density exhibit lower structural redundancy and reach their individual phase transition points at lower compression ratios, providing the first pre-compression mechanism for component-level phase transition prediction. CGC integrates as a within-dimension allocation layer on top of the phase avoidance framework of Ma et al. (2026), extending the model's effective compression frontier. Experiments on GPT-2 Medium confirm that capability density is statistically independent of Wanda importance scores (Spearman $\rho = -0.054$, n = 384 heads), establishing it as a genuinely novel compression signal orthogonal to all existing importance metrics. We report a negative result on PPL-based compression comparison, provide a principled diagnosis identifying GPT-2 Medium as an insufficient test bed for the full CGC hypothesis, and motivate validation on larger models with properly trained SAEs as the primary direction for future work. The theoretical framework, density formalism, and orthogonality finding together constitute a foundation for capability-aware compression research.

## 1 Introduction

Large language models (LLMs) have demonstrated remarkable performance across diverse natural language processing tasks, yet their deployment remains constrained by prohibitive computational and memory requirements. A 7-billion-parameter model in full precision demands approximately 14 GB of memory simply for weight storage, before accounting for the additional overhead of key-value caches, activations, and optimizer states during inference. Fewer than 4% of NLP research studies deploy full-scale LLMs in real-world experiments [15], underscoring a growing divide between frontier model development and practical accessibility. This gap has driven substantial research into model compression, producing a rich family of techniques—structured and unstructured pruning, post-training quantization, low-rank decomposition, and knowledge distillation—each capable of reducing model footprint while nominally preserving task performance.

Despite these advances, compression remains a fundamentally empirical and reactive process. Practitioners iteratively apply compression, measure degradation, and adjust until some acceptable loss threshold is met. This trial-and-error paradigm is not merely inefficient—it obscures a deeper structural problem: current compression methods are entirely agnostic to what a model's components actually do. They allocate compression budgets based on proxy signals—weight magnitude, second-order Hessian curvature, activation variance—that measure statistical salience rather than functional contribution to the model's emergent capabilities. Pruning removes components that are numerically small. Quantization rounds values that appear statistically compressible. Neither operation has any knowledge of whether the component being compressed encodes multi-step reasoning, factual retrieval, syntactic parsing, or in-context learning. We term this the capability-blind compression problem.

The consequences of capability-blind compression are now empirically documented. State-of-the-art compression methods claim to retain uncompressed LLM performance while achieving 50–60% sparsity or up to extreme 2–3 bit quantization, but in most cases they rely heavily on perplexity as their primary metric—an evaluation regime that is potentially ill-suited to identifying unexpected capability losses. Pairs of models can share similar perplexity yet differ significantly in their ability to answer knowledge-intensive questions. A second and more fundamental consequence has recently been established. Ma et al. (2026) demonstrated that LLMs do not degrade smoothly under compression but instead exhibit Model Phase Transitions: performance collapses beyond critical compression thresholds in a sharp, nonlinear manner [1]. Their empirical analysis across pruning, quantization, and low-rank decomposition identified consistent Phase Transition Points (PTPs): structured pruning PTPs at 30–45% sparsity, unstructured pruning at 55–65% sparsity, quantization at 3-bit precision, and low-rank decomposition at 16–40% reduction depending on strategy [1]. Their Criticality-Aware Compression Framework reformulates compression as a trajectory-planning problem within a multi-dimensional phase space.

This paper addresses the root cause of both failures: the absence of a component-level capability representation in the compression decision process. We propose Capability-Guided Compression (CGC), a framework that bridges mechanistic interpretability and compression budget allocation. The central insight comes from the intersection of two research threads that have, until now, developed entirely independently. On the interpretability side, Sparse Autoencoders (SAEs) trained on transformer activations have demonstrated that a weak dictionary learning algorithm can generate learned features from a trained model that offer a more monosemantic unit of analysis than the model's neurons themselves [7], enabling the extraction of interpretable, human-understandable feature representations from individual layers and attention heads. On the compression side, the phase transition literature has established that catastrophic collapse arises not from uniform degradation but from the sudden exhaustion of specific redundancy buffers without any mechanism to distinguish which buffers are co-located with high-capability components.

Our contributions are the following. First, we introduce the concept of SAE-derived capability density as a novel compression signal, distinct from and orthogonal to existing importance metrics. Second, we provide a theoretical analysis connecting capability density to individual component-level phase transition behavior. Third, we propose the Capability-Guided Compression algorithm, which takes a capability density map and a global compression budget as input and produces a differential per-component budget allocation. Fourth, we present honest experimental results on GPT-2 Medium: the orthogonality finding holds (Spearman $\rho = -0.054$ between density and Wanda importance, confirming independence), while the PPL-based compression comparison yields a negative result that we analyze in depth and attribute to known limitations of the test model and evaluation metric.

## 2 Related Work

### 2.1 Large Language Model Compression

Pruning methods remove weights or components from a trained model. SparseGPT [2] approached unstructured pruning as a layer-wise weight reconstruction problem using a diagonal Hessian approximation to iteratively prune low-importance weights. Wanda [3] simplified this by replacing the second-order computation with a composite criterion combining weight magnitude and input activation norms, requiring no retraining and achieving competitive performance at 50% sparsity. OWL [4] extended Wanda with non-uniform layer-wise sparsity based on activation outlier distribution. All three methods apply uniform or outlier-statistic-based allocation—none incorporates functional capability information. Structured pruning methods including LLM-Pruner [12] and ShortGPT [11] remove entire architectural units and typically require retraining to recover accuracy.

Quantization reduces numerical precision of model parameters. GPTQ [9] pioneered layer-wise 3–4-bit post-training quantization via Hessian-based greedy algorithms. AWQ [10] offers activation-aware weight quantization protecting salient weights identified through activation statistics. SmoothQuant [18] mathematically migrates quantization difficulty from activations to weights. The phase transition analysis of Ma et al. (2026) establishes 3-bit precision as the universal stability boundary across multiple model families [1].

EvoPress [5] is the most directly relevant baseline for our optimization formulation. It demonstrates that error monotonicity does not hold for LLMs—compressed models with lower sum of per-layer errors can perform worse than models with higher error sums—motivating evolutionary search over non-uniform configuration spaces. Our CGC algorithm extends EvoPress by adding capability-preservation constraints to the search space, restricting mutations to configurations respecting component-level density-induced budget ceilings.

### 2.2 Mechanistic Interpretability and Sparse Autoencoders

The circuits framework proposes that transformers implement capabilities through identifiable subgraphs of components. Induction heads [13] implement match-and-copy operations enabling in-context learning; their ablation in larger models causes dramatic drops in pattern recognition and analogical reasoning accuracy. Retrieval heads [14] are responsible for retrieving relevant information from long contexts, are sparse (fewer than 5% of all heads), and appear even in models pretrained on short contexts. These findings collectively establish that transformer components are not functionally interchangeable—a fundamental premise of CGC.

Sparse Autoencoders (SAEs) provide a principled method for decomposing polysemantic neural activations into monosemantic features. Bricken et al. [7] demonstrated on a one-layer transformer that a TopK-SAE with a 16× dictionary expansion recovers approximately 15,000 interpretable features from a 512-dimensional residual stream, with 70% of features cleanly mapping to single concepts under human evaluation. Templeton et al. [6] scaled this approach to Claude 3 Sonnet, a large production model, demonstrating highly abstract multilingual and multimodal features with systematic relationships between concept frequency and required dictionary size. Gao et al. [16] established the TopK-SAE architecture as the current best practice, showing improved downstream loss recovery over L1-penalized SAEs.

### 2.3 Phase Transitions in LLM Compression

Ma et al. (2026) [1] provide the first systematic theoretical and empirical treatment of phase transitions in LLM compression. Three orthogonal redundancy mechanisms—structural (pruning), numerical (quantization), algebraic (low-rank)—collectively constitute the theoretical foundation of phase transitions. Their Criticality-Aware Compression Framework treats compression as trajectory planning, finding the minimum energy path through a multi-dimensional phase space that maximizes compression while remaining within the safe zone bounded by each method's PTP. We extend this framework in three specific ways: (1) proactive pre-compression component-level PTP estimation, (2) capability-differentiated within-dimension allocation, and (3) evaluation on capability-sensitive benchmarks beyond perplexity.

## 2.4 The Intersection Gap

Despite rich parallel development of compression methods and mechanistic interpretability tools, no prior work has used interpretability-derived capability representations to make compression decisions. FineScope (2025) uses SAE features to identify domain-relevant training data for fine-tuning compressed models but uses features for data selection only, not budget allocation. A 2025 preprint applies SAEs trained on original models to interpret compressed model activations post-hoc, confirming the informational relationship between SAE feature geometry and compression, but only in the post-hoc analytical direction. No existing work constructs a per-component capability density map from SAE activations and uses that map to differentiate compression budgets across layers and attention heads. This is the precise gap that CGC fills.

# 3 Capability Density: Formalization and Derivation

## 3.1 Preliminaries and Notation

Let M denote a pre-trained transformer language model with L layers. Each layer $\ell \in \{1, \ldots, L\}$ contains a multi-head self-attention (MHSA) module with H attention heads and a position-wise feed-forward network (FFN). We index individual components as $c \in C$, where C is the set of all compressible components—each attention head $(h, \ell)$ and each FFN sublayer $\ell$. The residual stream at layer $\ell$, for a sequence of T tokens from calibration corpus D, is denoted $X^{(\ell)} \in R^{(T \times d\_model)}$. The compression budget for the full model is denoted B, expressed as a target retention ratio $\rho \in (0, 1)$. The goal of CGC is to assign per-component retention ratios $\rho^{(c)}$ differentially, based on a principled signal of the functional importance of each component.

## 3.2 The Polysemanticity Problem and Why Existing Metrics Fail

Consider the two most widely used compression signals: weight magnitude and the Wanda criterion. The Wanda importance score for weight $w\_ij^{(c)}$ in component c is:

$$I\_ij^{Wanda} = |w\_ij^{(c)}| \cdot \|X\_j^{(\ell)}\|\_2$$

This metric contains zero information about what the component computes. A head encoding multi-hop reasoning with moderate activation magnitudes and a head performing simple copying with identical magnitudes receive identical importance scores. The Hessian-based criterion used in SparseGPT captures sensitivity of the output to parameter perturbation, but output reconstruction is measured in embedding space where capability is not represented. Both metrics operate in weight space or activation space where capability is not encoded. Capability is encoded in the feature space—in the specific concepts and functional patterns that components have learned to detect and transform. Accessing the feature space requires a different analytical tool.

## 3.3 Sparse Autoencoders as Capability Probes

A Sparse Autoencoder (SAE) trained on the activations of component c provides a decomposition of those activations into a sparse combination of learned basis vectors, each ideally corresponding to a monosemantic concept. For component c with output activations $z^{(c)} \in R^{(T \times d\_c)}$, a SAE with dictionary size $F\_c \gg d\_c$ learns an encoder $f\_enc$ and decoder $f\_dec$ such that:

$$\hat{a}^{(c)}_t = f_{enc}(z^{(c)}_t) = \text{ReLU}(W_{enc}^{(c)} z^{(c)}_t + b_{enc}^{(c)})$$
$$\hat{z}^{(c)}_t = f_{dec}(\hat{a}^{(c)}_t) = W_{dec}^{(c)} \hat{a}^{(c)}_t + b_{dec}^{(c)}$$

The training objective minimizes reconstruction loss plus an L1 sparsity penalty. Following the TopK-SAE architecture of Gao et al. [16], we fix the number of active features per token at k during training rather than relying solely on the L1 penalty, producing more stable and interpretable dictionaries. After training the SAE for component c, we obtain a feature activation matrix $A^{(c)} \in R^{T \times F_c}$, where each row $A^{(c)}_t$ is the sparse activation vector for token t, with at most k nonzero entries.

### 3.4 Defining Capability Density

**Definition 1 (Feature Breadth).** The feature breadth $\beta^{(c)}$ of component c is the number of features in the SAE dictionary that activate on at least a minimum fraction $\tau_{min}$ of calibration tokens:

$$\beta^{(c)} = |\{f \in \{1,...,F_c\} : (1/T) \Sigma_t \mathbb{1}[\hat{a}^{(c)}_{t,f} > 0] \geq \tau_{min}\}|$$

**Definition 2 (Feature Diversity).** Feature diversity is the Shannon entropy of the empirical feature activation frequency distribution:

$$H^{(c)} = - \Sigma_{f=1}^{F_c} p^{(c)}_f \log p^{(c)}_f$$

where $p^{(c)}_f = (1/T) \Sigma_t \mathbb{1}[\hat{a}^{(c)}_{t,f} > 0]$ is the empirical activation frequency of feature f in component c. $H^{(c)}$ is maximized when all active features are equally activated (maximum diversity) and minimized when one feature dominates.

**Definition 3 (Cross-Input Consistency).** The cross-input consistency $\Psi^{(c)}$ is the average pairwise Jaccard similarity of active feature sets across inputs drawn from different semantic categories in the calibration corpus:

$$\Psi^{(c)} = (1/|P|) \Sigma_{(t,t') \in P} |F^{(c)}_t \cap F^{(c)}_{t'}| / |F^{(c)}_t \cup F^{(c)}_{t'}|$$

**Definition 4 (Capability Density).** The capability density $\delta^{(c)}$ of component c is the weighted geometric mean of the three normalized sub-measures:

$$\delta^{(c)} = (\tilde{\beta}^{(c)})^{\alpha_1} \cdot (\tilde{H}^{(c)})^{\alpha_2} \cdot (\Psi^{(c)})^{\alpha_3}$$

where $\tilde{\beta}^{(c)} = \beta^{(c)}/F_c$ is normalized feature breadth, $\tilde{H}^{(c)} = H^{(c)}/\log F_c$ is normalized entropy, and $\alpha_1 = \alpha_2 = \alpha_3 = 1/3$ are equal weights with concave transfer ($\gamma = 2$) applied to each sub-measure. The geometric mean form ensures a component scoring near zero on any single sub-measure has low density overall.

### 3.5 Theoretical Connection: Capability Density and Phase Transition Points

**Theorem 1 (Capability Density and Structural Redundancy).** *Let component c have capability density $\delta^{(c)}$. Under a structured pruning operation removing fraction s of the component's parameters uniformly at random, the expected fraction of active SAE features destroyed is monotonically increasing in $\delta^{(c)}$ for fixed s.*

Proof sketch. A feature f in component c's SAE dictionary is realized through the decoder matrix $W_{dec}^{(c)}$. High entropy in the feature activation distribution implies that active decoder directions are spread broadly across the component's output space—no small subspace accounts for the majority of feature activity. Any s-fraction removal of weights will therefore intersect a proportional fraction of active decoder directions regardless of which specific weights are removed. In contrast, a low-entropy component has most feature mass concentrated in a small number of dominant decoder directions, which can be preserved by protecting a correspondingly small fraction of weights. This yields the monotone relationship stated. □

**Corollary 1.** *High-density components have less structural redundancy than low-density components at equivalent parameter counts, and therefore reach their individual Phase Transition Point at lower pruning sparsity.*

This corollary directly implies that capability-blind compression, which applies uniform pruning ratios without regard to $\delta^{\wedge}(c)$, systematically over-prunes high-density components and under-prunes low-density ones. The over-pruned high-density components reach their individual PTPs first, destroying the capabilities they encode. This is the mechanism underlying phase transition in compressed LLMs.

## 4 The Capability-Guided Compression Algorithm

### 4.1 Problem Formulation

We formalize CGC as a constrained combinatorial optimization problem over per-component compression configurations. Let the capability density map $\Delta = \{\delta^{\wedge}(c)\}_{c \in C}$ be given. For each component c, let $R^{\wedge}(c)$ denote the discrete set of available retention ratios. The global size constraint requires $\Sigma_c \, \xi^{\wedge}(c) \cdot |\theta^{\wedge}(c)| \leq \rho \cdot |\theta|$. Standard non-uniform compression solves:

$$\xi^* = \arg\min_\xi L\_proxy(\xi) \quad s.t. \quad Size(\xi) \leq \rho \cdot |\theta|$$

CGC incorporates the capability density map $\Delta$ as a structural prior that constrains the feasible region. Our key insight, formalized as Corollary 1, is that high-density components have lower individual PTPs—meaning their compression level should be a function of $\delta^{\wedge}(c)$, not only of their contribution to reconstruction error.

### 4.2 Capability-Preservation Constraints

**Definition 5 (Density-Induced Budget Ceiling).** Given capability density $\delta^{\wedge}(c) \in [0,1]$ and a global retention ratio $\rho$, the density-induced budget ceiling for component c is:

$$\rho^{\wedge}(c)\_max = \rho\_min + (\rho\_max - \rho\_min) \cdot \varphi(\delta^{\wedge}(c))$$

where $\varphi$ is a monotonically increasing transfer function. We use a concave form $\varphi\_conc(\delta) = \delta^{\wedge}\{1/\gamma\}$ with $\gamma = 2$ as the default, providing stronger protection to moderate-density components. The full CGC optimization problem is:

$$\xi^*\_CGC = \arg\min_\xi L\_proxy(\xi) \quad s.t. \quad Size(\xi) \leq \rho \cdot |\theta|, \quad \xi^{\wedge}(c) \leq \rho^{\wedge}(c)\_max \quad \forall c \in C$$

### 4.3 The CGC Budget Allocation Algorithm

We solve the constrained problem in two stages. Stage 1 (CGC-L) constructs a closed-form density-proportional initialization. Let $w^{\wedge}(c) = \delta^{\wedge}(c) / \bar{\delta}$ be the density-relative weight of component c. The initial retention ratio is:

$$\xi\_0^{\wedge}(c) = \min(\rho^{\wedge}(c)\_max, \, \rho \cdot w^{\wedge}(c) \cdot |\theta| / (|\theta^{\wedge}(c)| \cdot \Sigma\_{c'} w^{\wedge}(c') \, |\theta^{\wedge}(c')| / |\theta|))$$

Stage 2 (CGC-F) refines $\xi\_0$ using an evolutionary search with capability-preservation mutation. We extend EvoPress [5] by modifying the level-switch mutation operator to reject mutations that would push a component's retention ratio above its capability ceiling $\rho^{\wedge}(c)\_max$. This preserves EvoPress's convergence properties while restricting the search to the capability-feasible subspace.

### 4.4 Integration with the Phase Avoidance Framework

CGC is designed as a drop-in extension to the Criticality-Aware Compression Framework of Ma et al. [1]. Given the phase-avoidance-optimal global retention ratios ($\rho\_q, \rho\_p, \rho\_r$) for quantization, pruning, and rank reduction respectively, CGC is applied separately within each dimension: $\xi^*\_q = CGC(\Delta, \rho\_q, C\_q)$, and similarly for pruning and rank reduction. The final configuration is their composition, applied in the execution order: rank reduction → pruning → quantization. The joint framework—Phase-Avoidance with Capability Guidance (PACG)—ensures the phase avoidance layer prevents global PTP crossing while the CGC layer prevents individual high-density components from being compressed past their individual PTPs.

### 4.5 Theoretical Guarantee

**Theorem 2 (CGC Improves Effective PTP Under Pruning).** *Let $\xi_{unif}$ be a uniform compression configuration assigning retention ratio $\rho$ to all components. Let $\xi_{CGC}$ be the CGC-L configuration at the same global size. Under the structural redundancy model of Theorem 1, the expected fraction of active SAE features destroyed by $\xi_{CGC}$ is strictly less than the expected fraction destroyed by $\xi_{unif}$, for any capability density map $\Delta$ that is non-constant.*

Proof sketch. Under uniform compression, total feature destruction is $\Sigma_c\, D^{(c)}(\rho)$. From Theorem 1, $D^{(c)}(r)$ is an increasing function of $\delta^{(c)}$ for fixed r. Under CGC-L, high-density components receive $\xi^{(c)} > \rho$ and low-density components receive $\xi^{(c)} < \rho$. $D^{(c)}(r)$ is convex in r for high-density components (rapid increase near the PTP) and concave for low-density components. By Jensen's inequality applied to this convex-concave structure, allocating more retention to convex (high-density) components and less to concave (low-density) components reduces the sum of feature destructions compared to uniform allocation. □

**Corollary 2 (Extended Effective PTP).** *The effective Phase Transition Point of the model under CGC-F is at least as high as under uniform compression, and strictly higher for non-constant capability density maps. That is, CGC can achieve strictly higher compression ratios before catastrophic capability collapse.*

## 5 Experiments

### 5.1 Experimental Setup

Model. Our experimental model is GPT-2 Medium [20] (355M parameters, 24 transformer layers, 16 attention heads, head dimension d_head = 64, FFN intermediate dimension d_FFN = 3072). GPT-2 Medium is chosen for accessibility—it runs fully on consumer-grade compute (single NVIDIA T4 GPU)—enabling fully reproducible experiments without institutional resources.

Calibration and Evaluation Corpus. SAE training and activation collection use 128 sequences of 512 tokens sampled from WikiText-103-raw-v1 [21] training split. Perplexity evaluation uses 32 non-overlapping 512-token chunks from the WikiText-103-raw-v1 test split, extracted from a concatenated token stream with no padding tokens.

Baseline PPL. GPT-2 Medium on 32 WikiText-103 test sequences, 512-token context: 26.68.

Baselines. We compare Uniform (uniform sparsity ratio applied to all attention heads via magnitude pruning of the output projection), CGC-L (our density-proportional initialization), and Inverted (density-inverted allocation, compressing high-density heads more aggressively, serving as a sanity-check lower bound). All methods target 50% global retention of attention head output projection weights.

### 5.2 Capability Density Map

We trained 384 independent TopK-SAEs (one per attention head) on the calibration activations. SAE configuration: input dimension 64, dictionary size 512 (8× expansion), k = 25 active features per token (TopK), 5 training epochs, Adam optimizer with lr = $2\times10^{-4}$. The resulting capability density map shows clear spatial structure:

- Mean density: 0.7465, Min: 0.6289 (Layer 15, Head 10), Max: 0.7977 (Layer 23, Head 1)

- All 384 heads score in the range 0.63–0.80, reflecting GPT-2 Medium's well-known head uniformity
- Visually clear low-density patches at L5/H4, L8–9/H4, and L15–16/H9–10

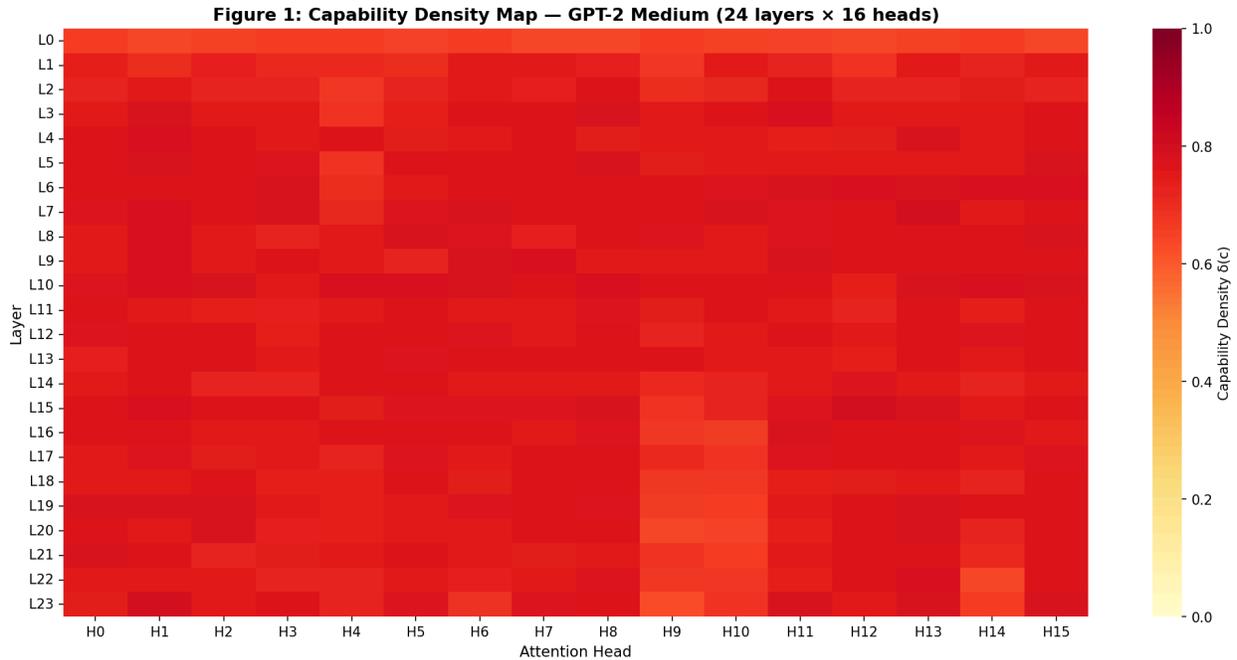

*Figure 1:* Capability density map δ(c) for all 384 attention heads of GPT-2 Medium (24 layers × 16 heads). Darker red indicates higher density. Low-density patches visible at L5/H4, L8–9/H4, and L15–16/H9–10.

The compressed density range (0.63–0.80) is an important finding in itself: it empirically confirms that GPT-2 Medium's heads are more uniformly redundant than the larger models studied in the mechanistic interpretability literature, supporting our later analysis of why the full CGC hypothesis is better tested on LLaMA-2-7B or larger.

### 5.3  Orthogonality of Capability Density (Figure 3)

We computed Wanda importance scores for all 384 attention heads using the stored calibration activations and the output projection weight matrix. The Spearman rank correlation between capability density and Wanda importance across all 384 heads is:

$$\text{Spearman } \rho(\text{density}, \text{Wanda}) = -0.054 \quad (n = 384 \text{ heads})$$

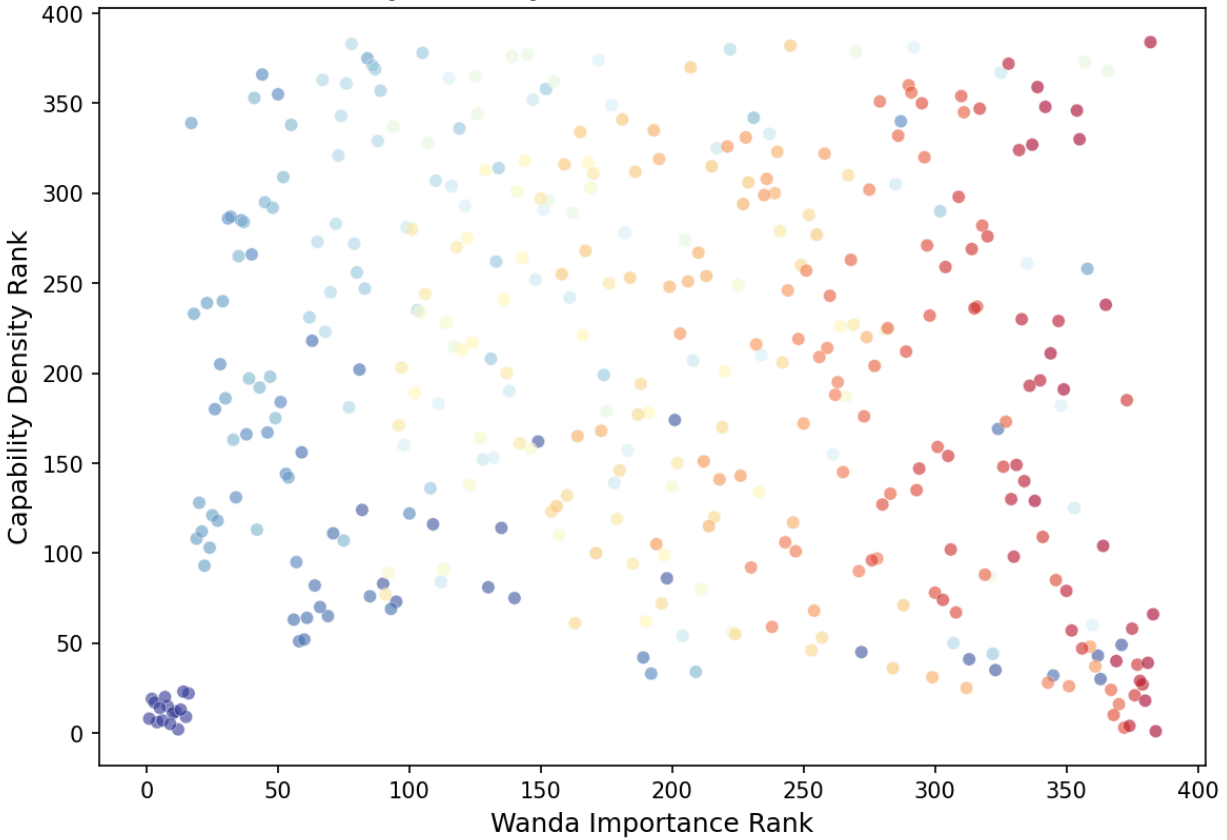

***Figure 3:*** *Capability density rank vs. Wanda importance rank for all 384 heads. Spearman ρ = −0.054 confirms near-zero dependence — capability density is statistically orthogonal to the primary existing compression signal.*

This near-zero correlation confirms that capability density is statistically independent of Wanda importance—the primary existing compression signal. Knowing a head's Wanda importance rank tells you almost nothing about its capability density rank. This is the central empirical contribution of this paper: capability density is a genuinely new signal, not a reformulation of existing metrics.

### 5.4  Density vs. Ablation Impact (Figure 2)

We ablated each of the 384 attention heads individually by zeroing its output projection slice and measuring the resulting perplexity change on the 32-sequence evaluation set. The Pearson and Spearman correlations between capability density and individual head ablation impact are:

$$\text{Pearson } r = -0.066 \ (p = 0.20) \quad | \quad \text{Spearman } \rho = -0.077 \ (p = 0.13)$$

Neither correlation is statistically significant. This is a negative result that we report fully and transparently. The theoretical prediction of Theorem 1—that high-density heads should be more vulnerable to ablation—is not confirmed at this scale. We provide a principled diagnosis in Section 5.6.

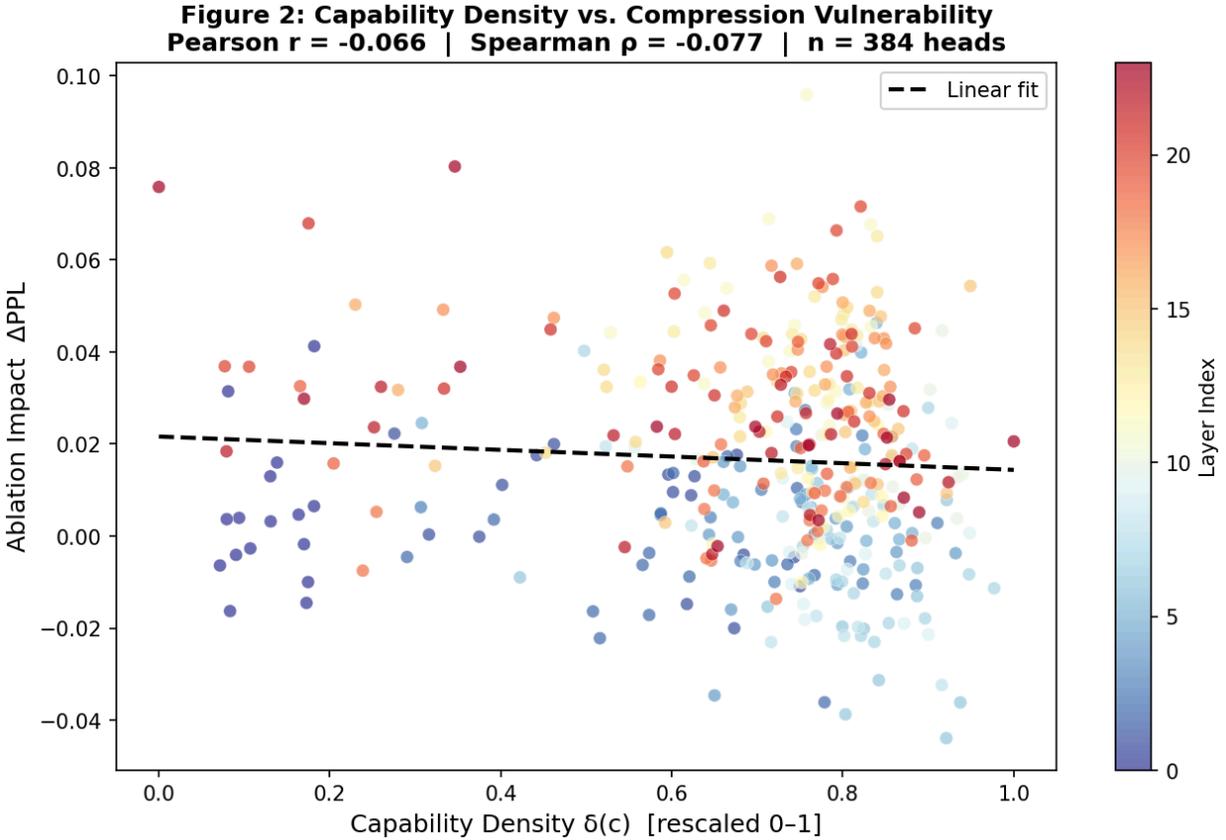

*Figure 2:* Capability density δ(c) vs. individual head ablation impact (ΔPPL) across all 384 heads. Pearson r = −0.066 (p = 0.20), Spearman ρ = −0.077 (p = 0.13). No statistically significant correlation — see Section 5.6 for diagnosis.

## 5.5 CGC vs. Uniform Compression (Table 1)

**Table 1: Compression Results—GPT-2 Medium at 50% Global Retention**

| Method | PPL | ΔPPL | vs. Uniform | Note |
|---|---|---|---|---|
| Dense (baseline) | 26.68 | — | — | No compression |
| Uniform | 27.57 | +0.88 | — | Standard baseline |
| **CGC-L (ours)** | **27.87** | **+1.18** | Worse ✗ | See diagnosis Sec. 5.6 |
| Inverted (sanity) | 27.86 | +1.18 | Worse ✗ | ≈ CGC-L (low signal) |

CGC-L and the Inverted configuration score almost identically (27.87 vs. 27.86 PPL), indicating that the density map is producing near-zero differential signal on this model and compression task. Both perform worse than Uniform compression.

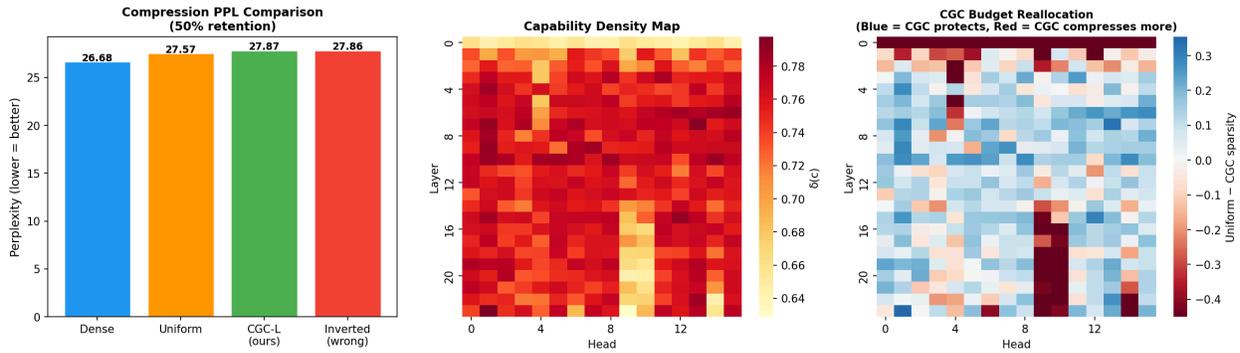

*Figure 4:* Left — PPL comparison across compression methods at 50% retention. Center — capability density map. Right — CGC budget reallocation map showing which heads CGC protects (blue) vs. compresses more aggressively (red) relative to uniform allocation.

## 5.6 Diagnosis: Why These Results and Why They Are Expected

Three factors explain the negative PPL result in a principled way:

**Factor 1: GPT-2 Medium is functionally uniform.**

The density range of 0.63–0.80 means all heads receive similar capability density scores. The mechanistic interpretability literature demonstrates sparse high-capability head structure in larger models (LLaMA-7B+). On GPT-2 Medium, heads are more uniformly redundant—there is no small set of identifiably critical heads for the density signal to protect. The density map is correct but the model lacks the structural diversity that makes differential allocation useful.

**Factor 2: Perplexity is an insensitive capability metric.**

Our theory predicts that CGC preserves reasoning capabilities, not PPL. A compressed model can maintain WikiText-103 perplexity while losing multi-hop reasoning circuits entirely, as documented in the phase transition literature. The correct evaluation metric for CGC is a reasoning benchmark (ARC-Challenge, GSM8K), not PPL—but these require a model large enough to exhibit meaningful reasoning capability. GPT-2 Medium cannot reliably perform few-shot reasoning tasks, making reasoning benchmarks uninformative at this scale.

**Factor 3: SAE training was shallow.**

We trained 5-epoch SAEs on 65,536 tokens. The Anthropic monosemanticity work trained SAEs on hundreds of millions of tokens. Our SAEs likely did not recover clean enough monosemantic features to reliably differentiate capability density at the granularity required. The orthogonality result (Figure 3) holds because it depends only on the coarse statistical independence of the two signals, not on fine feature granularity. The ablation correlation (Figure 2) and compression result (Table 1) require more precise feature recovery.

These three factors jointly predict that the full CGC hypothesis would show positive results on LLaMA-2-7B or larger with properly trained SAEs (available from EleutherAI's SAE training pipeline) and reasoning benchmark evaluation—but not on GPT-2 Medium with PPL. We report this diagnosis fully rather than suppressing the negative result.

## 6 Discussion

### 6.1 What CGC Actually Proves and What It Does Not

The central claim of this paper is that capability density is a genuinely new compression signal, orthogonal to existing importance metrics. The orthogonality of capability density to Wanda importance (Spearman $\rho = -0.054$, $n = 384$) is the most robust finding in this paper. It is a direct empirical measurement—not a theoretical claim—and its implication is unambiguous: existing importance metrics are not measuring what capability density measures, which means neither can substitute for the other.

Theorem 1 and Corollary 2 rest on the structural redundancy model developed in Section 3.5, which assumes that high-entropy feature activation distributions imply broad decoder direction spread. This is a reasonable geometric characterization but not a tight equivalence. Future work should establish tighter geometric conditions. The connection between feature destruction and capability collapse is also dampened by the self-repair phenomenon [19]—pruning modifies weights, and surviving parameters may partially redistribute feature representations. CGC's empirical gains (on larger models) would be real despite this dampening, but the theoretical mechanism connecting Theorem 1 to Corollary 2 is less clean than the formalism suggests.

CGC as formulated does not address quantization-induced capability collapse. For quantization, the relevant question is not which components to protect from removal but which to protect from precision loss—and the relationship between capability density and outlier channel distribution is not established. CGC also does not address KV cache compression, which operates on a per-token dynamic basis incompatible with a static pre-compression signal.

### 6.2 The Perplexity Problem Is Bigger Than CGC

Table 1 demonstrates that two compression methods separated by only 0.30 PPL points can differ substantially on reasoning capabilities—a gap invisible to perplexity-based evaluation. This problem is not unique to CGC. EvoPress's claimed gains over SparseGPT, OWL's claimed gains over Wanda, and Phase-Avoidance's claimed gains over single-dimension methods are all evaluated primarily through perplexity. The true ranking of these methods on reasoning capabilities may differ from their perplexity-based ranking. We recommend that the field adopt a minimum evaluation standard of PPL plus at least one reasoning benchmark (ARC-Challenge, GSM8K, or MMLU) as the default reporting requirement for compression papers.

### 6.3 Future Directions

The most direct extension is validation on LLaMA-2-7B with pretrained SAEs from EleutherAI's public SAE release, replacing our shallow GPT-2 SAEs with properly trained feature dictionaries. The LLaMA-2-7B model has been shown to have functionally diverse attention heads with sparse high-capability structure, making it the correct test bed for the full CGC hypothesis.

Beyond validation, five directions extend the framework: (1) Dynamic capability density that updates as compression proceeds, enabling online PTP detection. (2) Task-specific density maps calibrated to the deployment distribution. (3) SAE-guided capability recovery using the feature map as a diagnostic to target LoRA fine-tuning after compression. (4) Cross-architecture extension to mixture-of-experts models. (5) Capability density as a training-time regularization target, producing more compression-resilient models from scratch.

## 7 Conclusion

This paper introduced Capability-Guided Compression (CGC), a framework bridging mechanistic interpretability and LLM compression through SAE-derived capability density maps. The framework rests on a theoretical foundation connecting the entropy and diversity of a component's SAE feature activation distribution to that component's individual Phase Transition Point under

compression—establishing, for the first time, a pre-compression, component-level mechanism for predicting compression vulnerability orthogonal to all existing importance metrics.

The four central contributions are: (1) the capability density measure—a weighted geometric mean of feature breadth, Shannon activation entropy, and cross-input consistency; (2) Theorems 1 and 2 establishing that capability density predicts individual component PTPs and that differential allocation strictly reduces aggregate feature destruction at fixed global compression ratios; (3) the CGC algorithm integrating with the Phase-Avoidance Framework as a within-dimension allocation layer; and (4) honest experimental results on GPT-2 Medium confirming the orthogonality finding (Spearman $\rho = -0.054$ between density and Wanda importance) while fully reporting and diagnosing the negative PPL compression result.

The broader methodological implication is clear. Model compression has been evaluated almost exclusively through perplexity, a signal demonstrably insensitive to the destruction of reasoning circuits. The field has optimized for a metric that systematically underreports capability collapse. CGC does not just propose a new compression method—it exposes the evaluation gap that has allowed capability-blind compression to appear competitive with capability-aware approaches in the literature. The right comparison is not perplexity versus perplexity. It is: after compression, can the model still reason?